\documentclass[letterpaper, 10 pt, conference]{ieeeconf}  

\usepackage{graphicx}
\usepackage{epstopdf}
\usepackage{subfigure}
\usepackage{times}
\usepackage{url}
\usepackage[bookmarks=true, bookmarksopen=true, bookmarksnumbered=true]{hyperref}
\hypersetup{hidelinks}
\usepackage[font=footnotesize, labelsep=colon]{caption}
\usepackage{bm}
\usepackage{amssymb}
\usepackage{amsmath}
\usepackage[ruled]{algorithm2e}
\usepackage{algorithmic,float}
\usepackage{multirow}
\usepackage{booktabs}
\usepackage{array}
\usepackage{threeparttable}

\IEEEoverridecommandlockouts                                         

\title{\LARGE \bf
3D Lidar Mapping Relative Accuracy Automatic Evaluation Algorithm
}

\author{Guibin Chen, Jiong Deng, Dongze Huang, Shuo Zhang
\thanks{All the authors are with the Alibaba DAMO Academy Autonomous Driving Lab, Hangzhou 311121, China. Corresponding author: Guibin Chen, email: {\tt\small guibin.cgb@alibaba-inc.com}}
}

\begin{document}

\maketitle
\thispagestyle{empty}
\pagestyle{empty}

\begin{abstract}

HD (High Definition) map based on 3D lidar plays a vital role in autonomous vehicle localization, planning, decision-making, perception, etc. Many 3D lidar mapping technologies related to SLAM (Simultaneous Localization and Mapping) are used in HD map construction to ensure its high accuracy. To evaluate the accuracy of 3D lidar mapping, the most common methods use ground truth of poses to calculate the error between estimated poses and ground truth, however it's usually so difficult to get the ground truth of poses in the actual lidar mapping for autonomous vehicle. In this paper, we proposed a relative accuracy evaluation algorithm that can automatically evaluate the accuracy of HD map built by 3D lidar mapping without ground truth. A method for detecting the degree of ghosting in point cloud map quantitatively is designed to reflect the accuracy indirectly, which takes advantage of the principle of light traveling in a straight line and the fact that light can not penetrate opaque objects. Our experimental results confirm that the proposed evaluation algorithm can automatically and efficiently detect the bad poses whose accuracy are less than the set threshold such as 0.1m, then calculate the bad poses percentage $\mathcal{P}_{bad}$ in all estimated poses to obtain the final accuracy metric $\mathcal{P}_{acc} = 1 - \mathcal{P}_{bad}$.

\end{abstract}

\section{\textsc{Introduction}}

In recent years, with the rise of AI (Artificial Intelligent), autonomous driving technology has been greatly developed. The HD map with centimeter level accuracy plays an important role in the localization, planning, decision-making and perception of autonomous vehicles, which is usually built by 3D lidar mapping algorithm. After the HD map has been constructed, we need to evaluate its accuracy to improve the related mapping algorithm or check whether the HD map meets the accuracy requirement of actual use.

\subsection{Problem Description}

As everyone knows, the process of mapping is that sequences of 3D point clouds are transformed from body coordinate system to world coordinate system according to their corresponding estimated poses, so evaluating the map accuracy is actually evaluating the estimated poses accuracy, which is based on the premise that all 3D point clouds have been removed the distortion caused by vehicle movement and lidar rotation through the motion compensation algorithm. The accuracy of pose is usually divided into absolute accuracy and relative accuracy, the former is usually not ambiguous, but the latter has different meanings in different papers. In this paper, the relative accuracy refers to the relative position accuracy between different poses, which is usually presented by the degree of ghosting in the point cloud map: the lower the relative accuracy of poses, the more serious the ghosting problem of the point cloud map, and the more unfavorable it is for online localization algorithm based on point cloud matching.

\begin{figure}[tpb]
   \centering
   \includegraphics[width=8.6cm,height=2.5cm]{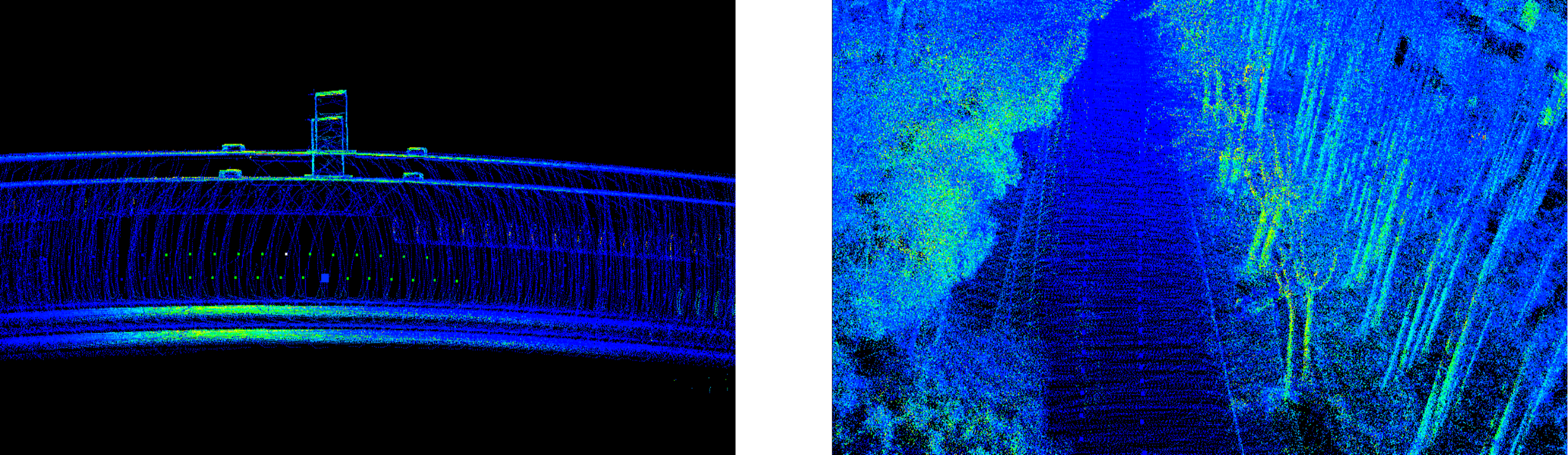}
   \caption{Serious ghosting phenomenon of point cloud map in tunnel and campus scenes. This problem is mainly caused by a large error in relative position between the different poses, and those poses are called bad poses with low relative accuracy in this paper.}
   \label{ghpts}
\end{figure}

In the field of autonomous driving, the accuracy evaluation metric of the HD map is usually expressed by the percentage $\mathcal{P}_{acc}$ : percentage of poses with relative accuracy greater than 0.1m in all estimated poses. The bad poses with relative accuracy less than 0.1m will cause serious ghosting in the corresponding areas of point cloud map as shown in Fig. \ref{ghpts}, which will gravely affect many functions based on HD map for autonomous vehicles, such as online localization. In this paper we also use the percentage $\mathcal{P}_{acc}$ as one of the output of our algorithm to describe the relative accuracy of 3D lidar mapping quantitatively.

The most common algorithms of map accuracy evaluation generally use ground truth of poses to calculate the error between estimated poses and ground truth, which belong to the field of absolute accuracy evaluation. But when the 3D lidar mapping algorithm is applied to the daily production of HD map, the ground truth of poses are usually not available, and to the best of authors knowledge, the 3D lidar mapping accuracy evaluation algorithm without ground truth is rarely studied in the academic and industry field which belongs to relative accuracy evaluation. The algorithm proposed in our paper is to solve this problem.

\subsection{Challenges}

Although we realize that when there is no ground truth of poses, we can only reflect the relative accuracy of the 3D point cloud map through the degree of possible ghosting in map, it's still not easy to automatically and efficiently detect and measure the ghosting. One of the challenges is the efficiency of the ghosting detection, nowadays, many HD map manufacturers check the accuracy of map by manual sampling inspection, but too few samples can not reflect the real situation of the map accuracy; too many samples will cost a lot of manpower or lead to low efficiency of map quality assessing. Another challenge is to detect the map ghosting along the three directions of XYZ axis. There is a method in the industry to evaluate the map relative accuracy by calculating the ground thickness in the point cloud map, but this method can only detect the ghosting along the Z axis (assuming that the Z axis is perpendicular to the ground). Similarly, the method of detecting the map ghosting along the XY axis through the thickness of traffic sign or lamp-post has theoretical defect: the thickness of traffic sign or lamp-post has no definite value which is different from the thickness of the ground.

\subsection{Related Works}

In order to assess the quality of mapping algorithms, the most common methods are to obtain the ground truth of poses, then calculate the APE (Absolute Pose Error) between estimate and reference \cite{grupp2017evo} \cite{zhang2018tutorial}, so, lots of related research works are focused on how to obtain the ground truth of poses. In \cite{geiger2012we} and \cite{salach2018accuracy}, the authors used the output of the GPS-IMU localization unit or the GNSS-RTK measurement as the ground truth for visual or lidar odometry/SLAM. However, these kinds of integrated navigation algorithms can't provide reliable ground truth poses in the area with poor satellite signal, which is caused by the shelter of tall buildings or trees. Based on the fact that lidar SLAM are usually 10 times more accurate than visual SLAM, \cite{caselitz2016monocular} obtained the ground truth camera trajectory composed by poses through a lidar-based SLAM system since the relative transformation between lidar and camera is known by calibration.

In addition to obtain the ground truth of poses, some works are dedicated to building the ground truth map by employing professional surveying and mapping device equipped with redundant sensors \cite{chen2019towards} \cite{chen2018accuracy}. Some accuracy evaluation algorithms need manual operation, for example, \cite{kummerle2009measuring} required manually matching point cloud observed by lidar to avoid point cloud registration failure.

However, all the above methods are aimed at establishing benchmarks, which are usually used to compare the results of different mapping algorithms or improve the researchers' mapping algorithms. As far as the authors know, few work studies accuracy evaluation without ground truth when mapping (especially for 3D lidar mapping) algorithms are applied to the actual production of HD map. In this case, we can only indirectly calculate the relative accuracy of the map by detecting the map's overlap or ghosting. \cite{filatov20172d} mainly selected three kinds of geometric feature metrics to determine the 2D map's quality without the ground truth of poses or map: the proportion of free and occupied cells, the number of the corners and the enclosed areas, but those metrics are not suitable for 3D lidar mapping in outdoor scenes because the 3D point clouds in these scenes are usually unevenly distributed and their geometry is often irregular. In \cite{chandran2008assessing}, the authors suppose that there are many planar structures in the environment and segment the 3D point cloud map into plane patches, then check whether the following two types of suspicious plane appear: the intersecting plane patches that don't represent corners and the parallel plane patches very close to each other, which indicates that the map at the suspicious plane appears ghosting, and its relative accuracy there is relatively low. However, \cite{chandran2008assessing} mainly measures the quality of 3D laser map in urban environment where many artificial planar structures such as buildings exist. Once there are only some pole-liked objects in some scenes, this algorithm would not work.

In contrast to all the above accuracy evaluation approaches, we focus on designing a more general 3D lidar mapping relative accuracy automatic evaluation algorithm without the ground truth of poses or map by assessing the degree of ghosting in 3D point cloud map, where the map's ghosting is actually caused by the translation or rotation relative error of the related estimated poses.

\section{\textsc{Problem Statement}}

The input of our designed accuracy evaluation method is a sequence of estimated poses set $\mathcal{T} = \{\bm{T}_i \mid i = 1, ..., n\}$ and 3D point clouds set $\mathcal{C} = \{\bm{C}_i \mid i = 1, ..., n\}$, where each pose $\bm{T}_i \in SE(3)$ is a transformation matrix estimated by 3D lidar mapping algorithm and each point cloud $\bm{C}_i$ is a set of 3D points $\{\bm{P}_j \mid j = 1, ..., m\}$ observed by multi-line 3D lidar such as Velodyne's HDL-32E and has been motion compensated to correct distortion. Each point $\bm{P}_j \in \mathbb{R}^5$ is a vector of its coordinate under body coordinate system (e.g. vehicle coordinate system) $P_j(x, y, z)$ plus its scanID and fireID $(s, f)$ which represents the vertical and horizontal firing order of 3D lidar's lasers respectively. Pose $\bm{T}_i$ and point cloud $\bm{C}_i$ have been synchronized in time, so they have a one-to-one correspondence.

Our proposed method outputs the indices set $\mathcal{S}_b$ of bad poses whose relative accuracy are less than the set threshold 0.1m, $\mathcal{S}_b \subseteq \{i \mid i = 0, ..., n - 1\}$. Our method also outputs the relative accuracy evaluation metric: percentage $\mathcal{P}_{acc} = 1 - \mathcal{P}_{bad}$, where $\mathcal{P}_{bad}$ is the percentage of bad poses in all input estimated poses. With an easy visualization tool based on PCL viewer, we can utilize the input and output data to view the ghosting situation of the point cloud submap around the bad poses, like the video we provided.

\section{\textsc{Proposed Method}}

\subsection{Map Ghosting Detection Principle}

Without the ground truth of poses or map, the intuitive way to judge the accuracy of a point cloud map constructed by 3D lidar mapping algorithm is to detect whether the ghosting appears in the map as in Fig. \ref{ghpts}. Our basic principle of detecting map ghosting is based on the fact that laser of 3D lidar travels in a straight line and can not penetrate opaque objects, which is often used for ray tracing, dynamic elements identification or estimating the quality of point cloud alignment by some researchers \cite{glassner1989introduction} \cite{pomerleau2014long} \cite{ding2019deepmapping}.

\begin{figure}[tbp] 
  \centering
  \includegraphics[scale=0.3]{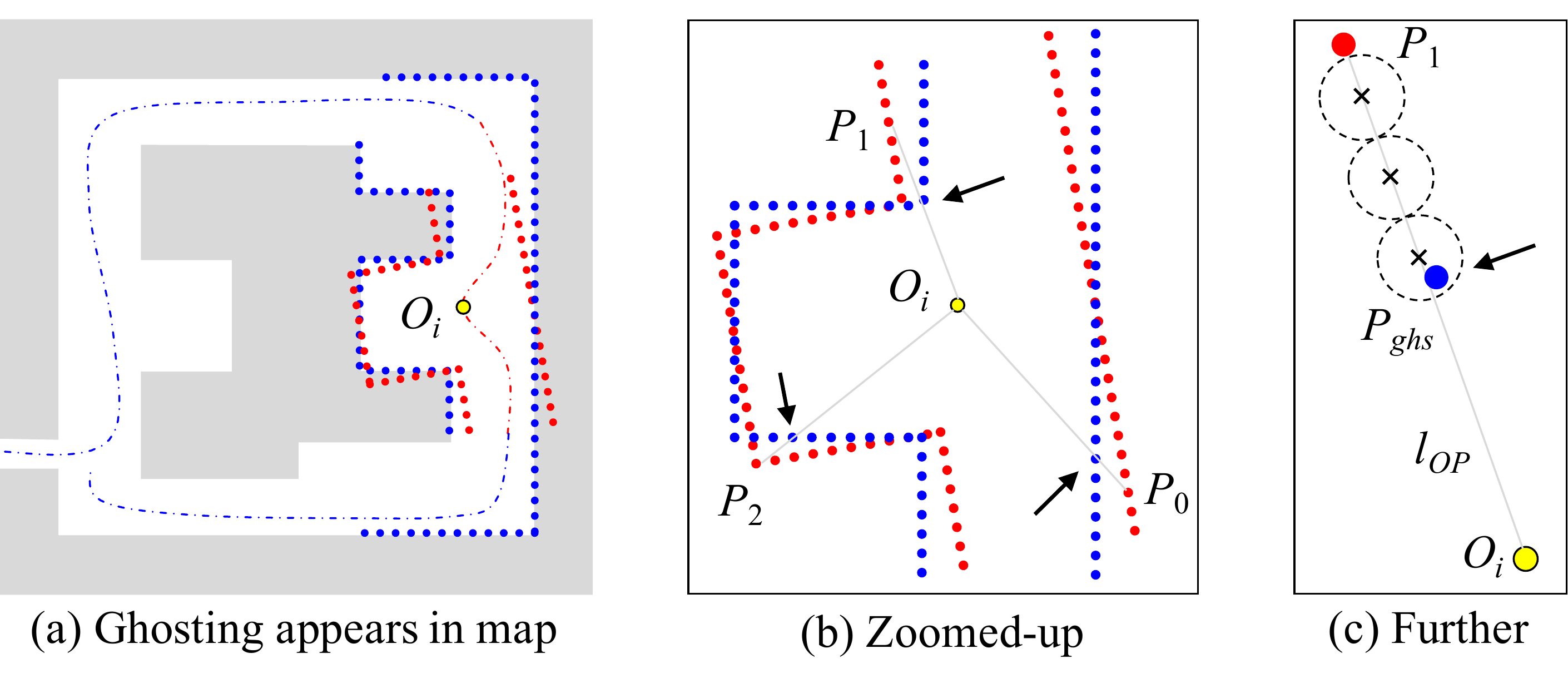} 
  \caption{An example of detecting ghosting in plan view. Red and blue points represent point cloud $\bm{C}_i^W$ and submap $\bm{C}_s^W$ respectively and we have not shown the ground points for display convenience. Dash-dotted line is trajectory composed by poses, where the red part represents the poses near the pose $\bm{T}_i$. In (b) (c), the gray solid line denote the line $l_{OP}$, and the black arrows emphasize the ghosting points. The black cross marks in (c) indicate the selected sampling positions and the black dashed circles around them represent the detection range $r_{ghs}$.}
  \label{ghdt}
\end{figure}

Fig. \ref{ghdt} shows an example of our method to detect ghosting: For pose $\bm{T}_i$, we first transform its corresponding point cloud $\bm{C}_i$ from body coordinate system to world coordinate system to get point cloud $\bm{C}_i^W$, then search for poses near the pose $\bm{T}_i$ with a certain radius $r_s$, also transform their corresponding point clouds into world coordinate system to make up the submap $\bm{C}_s^W$. If submap $\bm{C}_s^W$ don't contain ghosting, then there should be no points of $\bm{C}_s^W$ on line $l_{OP}$ from the lidar observation center $O_i$ to every point $P_j$ of $\bm{C}_i^W$. If not, ghosting point $P_{ghs}$ is detected and we call that the ghosting point $P_{ghs}$ is captured by the point $P_j$.

In order to discretize the above-mentioned ghosting detection process to facilitate programming, we select some sampling positions on line $l_{OP}$ starting from the end of point $P_j$, and only detect ghosting points at these positions, as shown in Fig. \ref{ghdt}(c).

Obviously, the proposed principle of ghosting detection will be broken by dynamic obstacles, so we will use some 3D obstacle perception algorithms \cite{li2016vehicle} \cite{wu2018squeezeseg} to remove them.

\subsection{Metric of Ghosting in Point Cloud Map}

In order to calculate the relative accuracy of pose $\bm{T}_i$, we first need to quantitatively describe the blur degree of ghosting detected in 3D point cloud map, where the required lidar observation center $O_i$ can be obtained by pose $\bm{T}_i$ and the calibration parameters from body coordinate system to lidar coordinate system.

\begin{figure}[bp] 
  \centering 
  \includegraphics[scale=0.3]{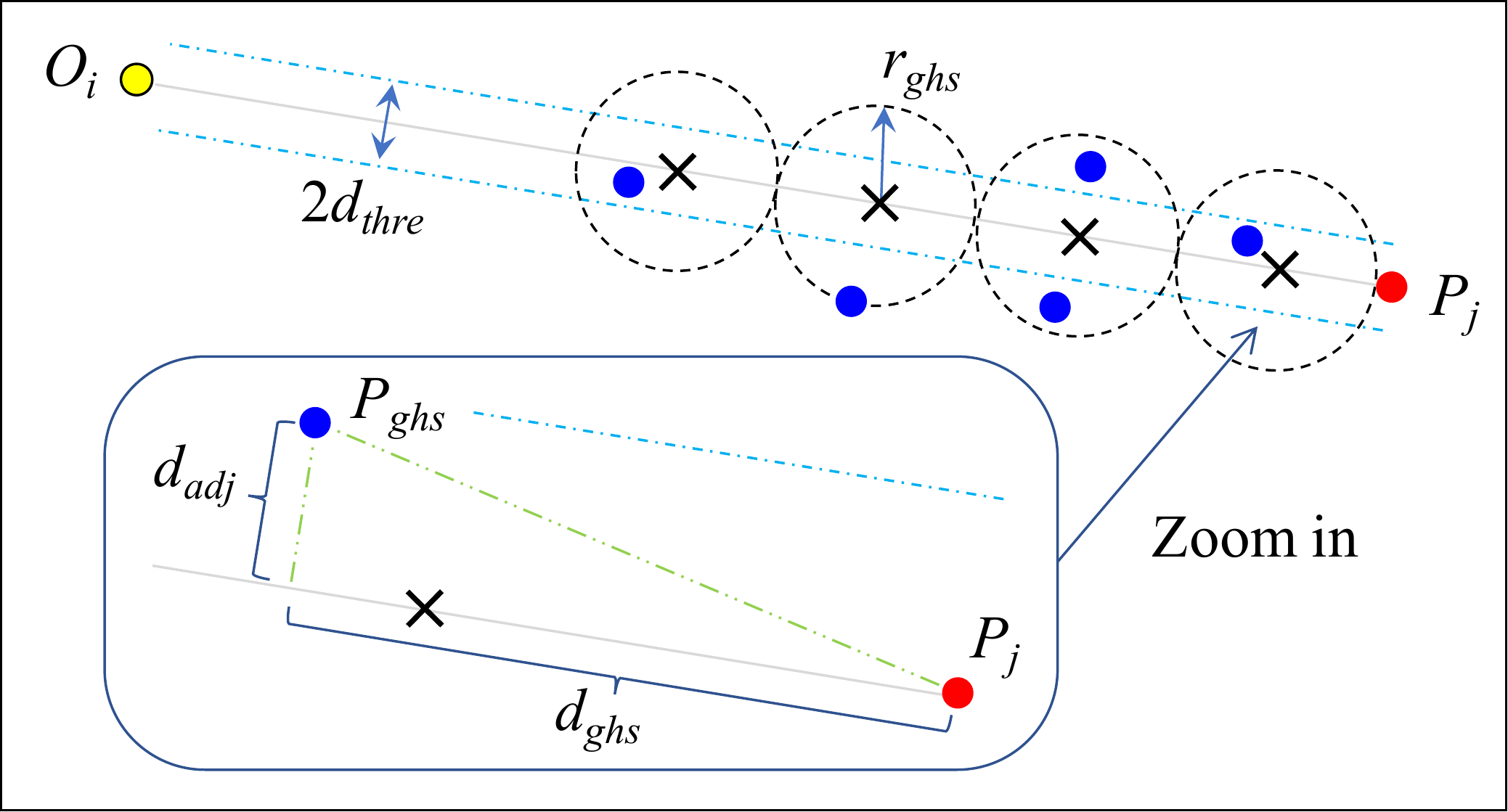} 
  \caption{Illustration of the basic quantized metric $d_{ghs}$ of ghosting. The two cyan dash-dotted lines show the tolerance of judging whether the ghosting point $P_ {ghs}$ is located on line $l_ {OP}$, and the other marks are same to Fig. \ref{ghdt}.}
  \label{ghme}
\end{figure}

Considering the measurement error ($1\sim2cm$, usually) of the 3D lidar sensor, we set a threshold $d_{thre}$ to determine whether the ghosting point $P_{ghs}$ is located on line $l_{OP}$, and the ghosting detection range $r_{ghs}$ should be slightly larger than the threshold $d_{thre}$. The basic quantized metric $d_{ghs}$ of ghosting is shown in Fig. \ref{ghme}.

To improve the search efficiency, we build the submap $\bm{C}_s^W$ into a kd-tree, and use its radius search function to search for ghosting points within radius $r_{ghs}$ at each sampling location. Once searched, the euclidean distance $d_{adj}$ is calculated by 

\begin{equation}
  \label{eq:d_adj}
  d_{adj} = \frac{| \overrightarrow{O_iP_j} \times \overrightarrow{P_{ghs}P_j} |} {| \overrightarrow{O_iP_j} |}
\end{equation}

If $d_{adj} < d_{thre}$, we determine that the searched ghosting point is located on line $l_{OP}$, so the basic quantized metric $d_{ghs}$ of ghosting given by

\begin{equation}
  \label{eq:d_ghs}
  d_{ghs} = \frac{\overrightarrow{O_iP_j} \cdot \overrightarrow{P_{ghs}P_j}}{| \overrightarrow{O_iP_j} |}
\end{equation}
is valid , and vice versa.

\begin{figure}[bp] 
  \centering 
  \includegraphics[scale=0.3]{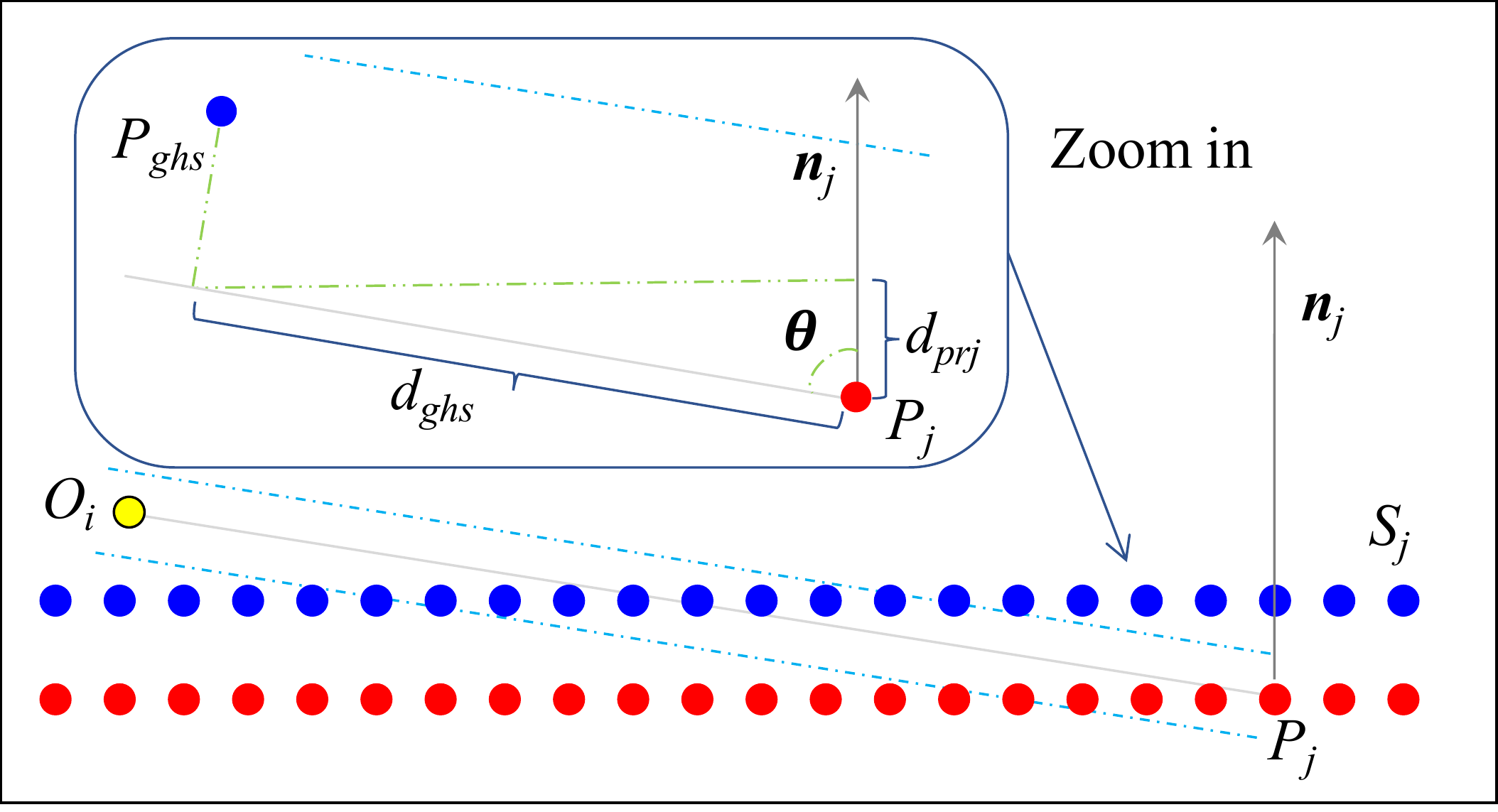} 
  \caption{Front view of the ground point cloud used to illustrate the more precise metric $d_{prj}$ of ghosting. The Blue and red points represent the ground in submap $\bm{C}_s^W$ and point cloud $\bm{C}_i^W$ respectively, and the brown up arrow above point $P_j$ indicates the normal vector of the ground surface.}
  \label{ghprj}
\end{figure}

However, the above basic metric $d_{ghs}$ can not exactly measure the ghosting appeared in point cloud map when line $l_{OP}$ is nearly parallel to the surface $S_j$ where the point $P_j$ is located, as shown in Fig. \ref{ghprj}. Thus, we must estimate the normal vector $\bm{n}_j$ of surface $S_j$ to get the more precise metric $d_{prj}$ of ghosting to solve this problem.

Using the established kd-tree of submap $\bm{C}_s^W$, we can easily obtain the point cloud $\bm{C}_j^s$ in the vicinity of surface $S_j$ and then calculate the normal vector $\bm{n}_j$ by PCA (Principal Component Analysis), mainly as follows:
\begin{equation}
  \begin{split}
  &\overline{P} = \frac{1}{k}\sum_{i=1}^{k}P_i, \forall{P_i}(x, y, z) \in \bm{C}_j^s \\
  &\bm{A}_{cor} = \frac{1}{k}\sum_{i=1}^{k}(P_i-\overline{P})(P_i-\overline{P})^T, \bm{A}_{cor} \in \mathbb{R}^{3\times3} \\
  &\bm{A}_{cor}\bm{V} = \bm{\lambda} \bm{V}, \text{solved by SVD, }\bm{\lambda}(\lambda_0<\lambda_1<\lambda_2)
  \end{split}
\end{equation}
where the eigenvalue $\lambda_0$ of the covariance matrix $\bm{A}_{cor}$ is significantly smaller than the other two eigenvalues $\lambda_1,\lambda_2$, since the point cloud $\bm{C}_j^s$ is approximately distributed on a planar patch. The eigenvector $\bm{v}_0$ in $\bm{V}(\bm{v}_0,\bm{v}_1,\bm{v}_2)$ corresponding to the minimum eigenvalue $\lambda_0$ denotes the normal vector of the planar patch: $\bm{n}_j = \bm{v}_0$.

Because the above problem only occurs when line $l_{OP}$ is approximately perpendicular to normal vector $\bm{n}_j$ of surface $S_j$, so we calculate the angle $\theta$ between $l_{OP}$ and $\bm{n}_j$ by
\begin{equation}
  \theta = arccos{\frac{\overrightarrow{O_iP_j} \cdot \bm{n}_j}{|\overrightarrow{O_iP_j}||\bm{n}_j|}}
\end{equation}
and combine $d_{adj}$ to switch the final metric $d_{prj}$ of ghosting
\begin{equation}
  d_{prj} = \begin{cases}
  d_{ghs} * \cos{\theta}, & \text{if, } d_{adj} < d_{thre} \\
  & \text{and, } \theta > \theta_{thre} \\
  d_{ghs}, & \text{elseif, } d_{adj} < d_{thre} \\ 
  0, & \text{otherwise}
  \end{cases}
\end{equation}

If we perform the above-mentioned ghosting detection on each point of the point cloud $\bm{C}_i^W$ and measure the detected ghosting points by the metric $d_{prj}$, the time-consuming of our entire 3D lidar mapping relative accuracy automatic evaluation algorithm will be terrible, because the original point cloud $\bm{C}_i$ of point cloud $\bm{C}_i^W$ usually contains tens of thousands of points, so we need to downsample the point cloud $\bm{C}_i$ and submap $\bm{C}_s^W$ through some certain strategies.

\subsection{Point Cloud Downsampling Method}

The main problem with submap $\bm{C}_s^W$ is that the point cloud is generally very dense due to the superposition of multiple point clouds, so we downsample it directly by the voxel grid filter with leaf size 0.02m.

However, the downsampling process of point cloud $\bm{C}_i$ is relatively complicated. On the one hand, the spatial position of the points in $\bm{C}_i$ can not be changed during the downsampling, otherwise the basic principle of ghosting detection will be broken; On the other hand, the spatial distribution of a single-frame laser point cloud like $\bm{C}_i$ is very uneven, and the ground points occupy most of the points near the lidar observation center $O_i$, but we do not want too many ground points to participate in ghosting detection; In addition, we hope to keep all the pole-liked points during the downsampling, which will facilitate ghosting detection in the scene where buildings are missing and only some trees.

So first, we extract the ground and pole-liked points from the point cloud $\bm{C}_i$ in body coordinate system through semantic segmentation of point cloud based on deep learning \cite{qi2017pointnet++}, and set the corresponding points label to ``ground" and ``pole", and the other points label to ``default".

\begin{figure}[bp]
   \centering
   \subfigure[before downsampling]{
   \includegraphics[width=4.05cm,height=2.6cm]{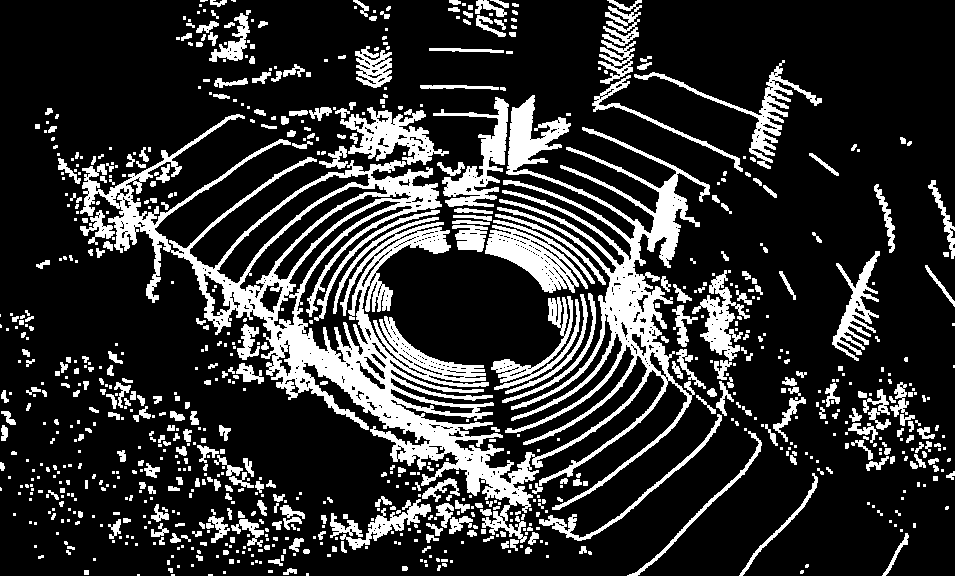}}
   \hspace{0in}
   \subfigure[after downsampling]{
   \includegraphics[width=4.05cm,height=2.6cm]{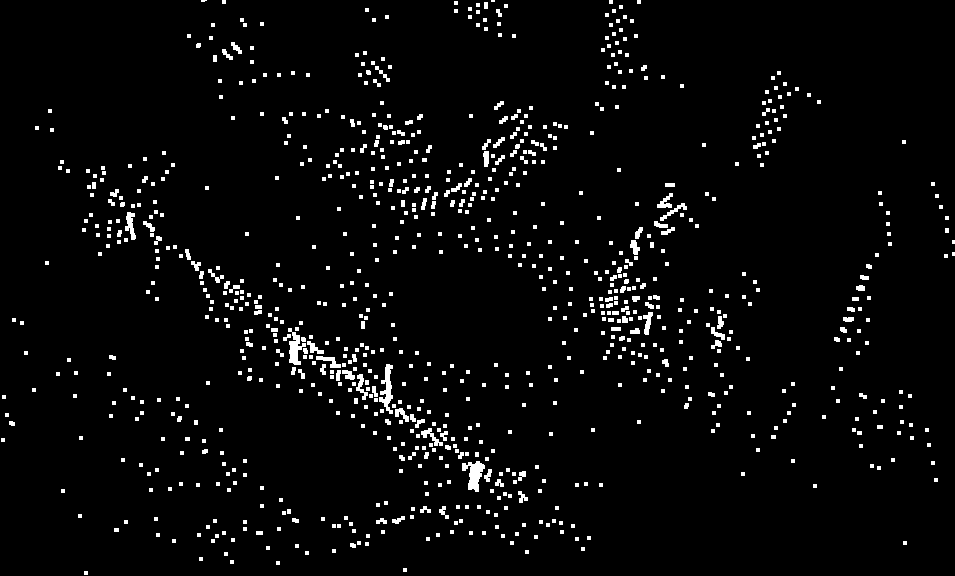}}
   \caption{The effect of downsampling for single-frame point cloud $\bm{C}_i$.}
   \label{fig:ds-effect}
\end{figure}

After obtaining the simple semantic information, we use the scanID and fireID of points in $\bm{C}_i$ to solve the problem that point cloud is dense in the middle and sparse in the outer, and finally realize the relatively uniform downsampling in space for point cloud $\bm{C}_i$ through the method summarized in Alg. \ref{alg:ds}. The effect is shown in Fig. \ref{fig:ds-effect}.

\begin{algorithm}[htbp]
\algsetup{linenosize=\footnotesize}
\footnotesize
\caption{downsampling for point cloud $\bm{C}_i$}
\label{alg:ds}
\KwIn{original point cloud $\bm{C}_i$}
\KwOut{downsampled point cloud $\bm{C}_i^d$}
Every Point $P \in \bm{C}_i$ has members $(x, y, z, s, f, l)$, which represent its coordinate, scanID, fireID, label, respectively.\\
$n_{l}$ is the lasers number of the 3D lidar sensor.\\
$\theta_{r}$ is the horizontal angular resolution($^{\circ}$) of lidar.\\
Let $f_{m} = 360 / \theta_{r}$, and $\xi = f_{m} / n_{l}$.\\
So $P(s)$ range $0, 1, ..., n_l - 1$; $P(f)$ range $0, 1, ..., f_{m} - 1$.\\
Set $\eta_{ground} = 30/\theta_{r}$.\\
Set $\eta_{vec} = \{\{5,6/\theta_{r}\},\{10,4/\theta_{r}\},\{20,2/\theta_{r}\},\{900,1/\theta_{r}\}\}$.\\
Init $\bm{C}_i^d$ empty.\\
\ForEach{$P \in \bm{C}_i$}{
  \eIf{$P(l)$ = ``pole"}{
    push $P$ to $\bm{C}_i^d$
  }{
    $f_{scat}=P(f) + P(s) * \xi$\\
    \If{$f_{scat} \ge f_{m}$}{
      $f_{scat} = f_{scat} - f_{m}$
    }
    \eIf{$P(l)$ = ``ground" and $f_{scat}\%\eta_{ground} = 0$}{
      push $P$ to $\bm{C}_i^d$
    }{
      \ForEach{$\eta \in \eta_{vec}$} {
        \If{$\|P(x,y,z)\| < \eta.first$ and $f_{scat}\%\eta.second = 0$} {
          push $P$ to $\bm{C}_i^d$\\
          \textbf{break}
        }
      }
    }
  }
}
\textbf{return} $\bm{C}_i^d$
\end{algorithm}

\subsection{Calculate the Relative Accuracy of Pose}
Using the downsampled submap $\bm{C}_s^W$ and point cloud $\bm{C}_i^W = \bm{T}_i\bm{C}_i^d$, we run the ghosting detection method and will get a series of ghosting points with metric $d_{prj}$. In order to determine whether the relative accuracy of pose $\bm{T}_i$ reaches 0.1m, we first need to record some statistics:
\begin{itemize}
  \item the labeled ``pole" points number $n_{pole}$ in the downsampled point cloud $\bm{C}_i^W$;
  \item the total number $n_{ordi}$ of the labeled ``ground" or ``default" points in $\bm{C}_i^W$;
  \item the number $m_{pole}$ of the ghosting points with metric $d_{prj}>0.1m$ which are captured by the labeled ``pole" points in $\bm{C}_i^W$;
  \item the number $m_{ordi}$ of the ghosting points with metric $d_{prj}>0.1m$ which are captured by the labeled ``ground" or ``default" points in $\bm{C}_i^W$.
\end{itemize}
Then, if $m_{pole}/n_{pole} > t_{pole}$ or $m_{ordi}/n_{ordi} > t_{ordi}$ where $t_{pole}$ and $t_{ordi}$ is the set threshold, we can determine that the pose $\bm{T}_i$ is a bad pose and insert the pose index $i$ to the bad poses indices set $\mathcal{S}_b$.

After completing the relative accuracy evaluation to all the estimated poses in $\mathcal{T}$, our method outputs the indices set $\mathcal{S}_b$ and calculates the percentage $\mathcal{P}_{bad}$ of the detected bad poses in $\mathcal{T}$ to present the relative accuracy evaluation metric: $\mathcal{P}_{acc} = 1 - \mathcal{P}_{bad}$.

\section{Experiments}

Our method evaluates the 3D lidar mapping relative accuracy through detecting the bad poses whose relative accuracy are less than 0.1m in the trajectory produced by the mapping algorithm and calculating their percentage in all estimated poses. However, all detection algorithms have a trade-off between the precision and recall rate. For the map accuracy evaluation problem in this paper, we are more inclined to have a higher recall rate of bad poses detection, and we achieved this by tuning some threshold parameters in the proposed accuracy evaluation algorithm.

In order to test the practical application effect of our algorithm, we need to experimentally verify the precision and recall rate of the bad poses detection. To this end, we first obtain ground truth poses in four different real-world scenes by running a 3D lidar mapping algorithm, which is based on LOAM \cite{zhang2014loam} and uses the output of integrated navigation device to perform a loop detection in the back end. After offline pose graph optimization and some necessary manual correction, the relative accuracy of ground truth poses can reach more than 0.03m.

\subsection{Precision Rate Verification}

The precision verification experiment is relatively simple: we directly apply our method to evaluate the relative accuracy of the ground truth poses in the four benchmark scenes. Because the ground truth poses have extremely high relative accuracy, once our method detects bad pose, it is a false detection, therefore, the metric $\mathcal{P}_{acc}$ output by our method is exactly the precision rate that needs to be verified in this experiment. The result is shown in Tab. \ref{tab:pre} which indicates that the precision rate of our designed relative accuracy evaluation algorithm is greater than 98\%.

\begin{table}[bp]
  \centering
  \caption{Result of precision rate verification}
    \begin{tabular}{ccccc}
    \toprule
    \multicolumn{1}{p{4.165em}<{\centering}}{Scene sequence} & \multicolumn{1}{p{4em}<{\centering}}{Distance (km)} & \multicolumn{1}{p{3.915em}<{\centering}}{Number of poses} & \multicolumn{1}{p{4.415em}<{\centering}}{Time cost (min)} & \multicolumn{1}{p{4.25em}<{\centering}}{Precision (\%)} \\
    \midrule
    0     & 5.24  & 4580  & 24    & 99.13 \\
    \midrule
    1     & 13.7  & 16818 & 45    & 99.48 \\
    \midrule
    2     & 22.8  & 11466 & 63    & 98.12 \\
    \midrule
    3     & 7.2   & 6325  & 29    & 98.03 \\
    \bottomrule
    \end{tabular}
    \begin{tablenotes}
       \footnotesize
       \item The used hardware condition: an Intel(R) Core(TM) i7-8700 CPU with 6 cores @ 3.20GHz with 32 GB RAM, and a Nvidia GeForce GTX 1080 with 8 GB RAM. And the time cost were tested on 10 threads.
    \end{tablenotes}
  \label{tab:pre}
\end{table}

\subsection{Recall Rate Verification}

In order to verify the recall rate of our method to detect bad poses, we need to artificially add some disturbances to the ground truth poses to intentionally generate ghosting in specified areas on the point cloud map. Meanwhile, we add disturbances to the $xy$ or $z$ value of ground truth poses separately to prove that our method can detect both the horizontal and vertical ghosting distributed along the XY and Z axis.

\begin{figure}[tbp] 
  \centering 
  \includegraphics[scale=0.3]{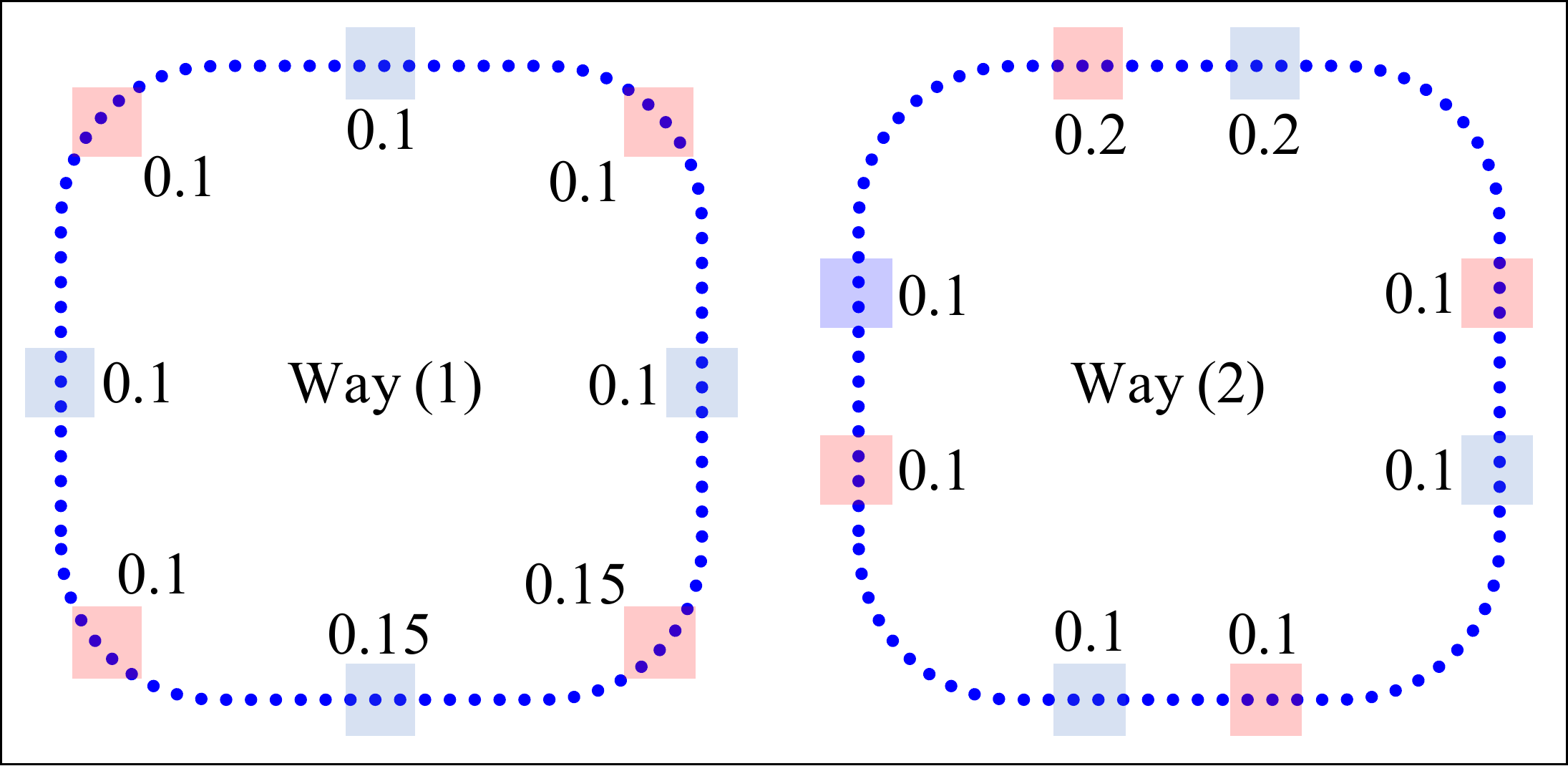} 
  \caption{The specific way of adding disturbances to the ground truth poses. Blue dotted line is the trajectory, and the upper red or blue rectangles indicate the areas where disturbances are added in the XY or Z direction. Numbers near the rectangles represent the magnitude of the disturbances.}
  \label{fig:distur}
\end{figure}

Fig. \ref{fig:distur} shows the specific way to add the disturbances: artificially add 0.1, 0.15, 0.2m disturbances to the $xy$ value or $z$ value of ground truth poses in specified areas of the trajectory according to the proportion of $6:1:1$, and the length of each disturbance is 50m.

We perform our accuracy evaluation method in the four different benchmark scenes where the artificially set disturbances are added through the way (1) and way (2) as illustrated in Fig. \ref{fig:distur}, and considering that one disturbed pose will cause ghosting in its surrounding point cloud submap, we use the area with disturbance as the statistical unit to calculate the recall rate.

\begin{figure*}[htbp]
  \centering 
  \includegraphics[scale=0.58]{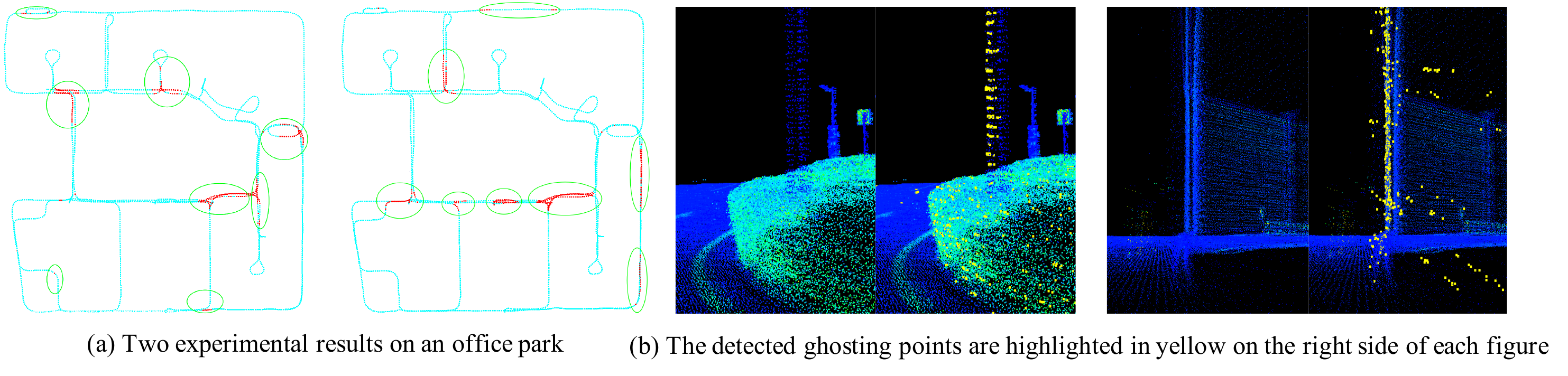} 
  \caption{Visualization results of partial recall rate experiments. (a) shows the trajectories of the accuracy evaluation results after adding disturbances according to the way (1) and way (2) illustrated in Fig. \ref{fig:distur}, where the red and cyan points represent the detected bad poses and the normal poses respectively, the green oval circles represent the ground truth of the perturbed areas. (b) shows the detected ghosting points which are colored yellow and located in the submap around some bad poses shown in (a).}
  \label{fig:trail}
\end{figure*}

The experimental result is shown in Tab. \ref{tab:rec}, and we can see that the recall rate for detecting the areas with disturbance of our proposed method is 100\%. This is mainly due to that we specially adjust some threshold parameters to improve the recall rate, which is actually at the expense of the precision rate. And Fig. \ref{fig:trail} shows the visualization results of the experiments executed in benchmark scene 0 which is an office park in the real world. More visualization information can be seen in the provided video.

\begin{table*}[htbp]
  \centering
  \caption{Result of recall rate verification}
    \begin{tabular}{ccccccccccc}
    \toprule
    \multicolumn{2}{c}{Direction of disturbance} & \multicolumn{1}{p{4.75em}<{\centering}}{XY} & \multicolumn{1}{p{4.75em}<{\centering}}{Z} & \multicolumn{1}{p{4.75em}<{\centering}}{XY} & \multicolumn{1}{p{4.75em}<{\centering}}{Z} & \multicolumn{1}{p{4.75em}<{\centering}}{XY} & \multicolumn{1}{p{4.75em}<{\centering}}{Z} & \multicolumn{1}{p{3.25em}<{\centering}}{XY} & \multicolumn{1}{p{3.25em}<{\centering}}{Z} & \multicolumn{1}{p{3.25em}<{\centering}}{XYZ} \\
    \midrule
    \multicolumn{1}{p{4.0em}<{\centering}}{Scene sequence} & \multicolumn{1}{p{4.0em}<{\centering}}{Distance (km)} & \multicolumn{2}{p{9.5em}<{\centering}}{Number of areas with 0.1m disturbance} & \multicolumn{2}{p{9.5em}<{\centering}}{Number of areas with 0.15m disturbance} & \multicolumn{2}{p{9.5em}<{\centering}}{Number of areas with 0.2m disturbance} & \multicolumn{2}{p{6.5em}<{\centering}}{Total number of all the areas} & \multicolumn{1}{p{3.25em}<{\centering}}{Recall (\%)} \\
    \midrule
    0     & 5.24  & 6     & 6     & 1     & 1     & 1     & 1     & \multirow{3}[8]{*}{32} & \multirow{3}[8]{*}{32} & \multirow{3}[8]{*}{100} \\
\cmidrule{1-8}    1     & 13.7  & 6     & 6     & 1     & 1     & 1     & 1     &       &       &  \\
\cmidrule{1-8}    2     & 22.8  & 6     & 6     & 1     & 1     & 1     & 1     &       &       &  \\
\cmidrule{1-8}    3     & 7.2   & 6     & 6     & 1     & 1     & 1     & 1     &       &       &  \\
    \bottomrule
    \end{tabular}%
  \label{tab:rec}%
\end{table*}%

\subsection{Discussion}

During evaluating the relative accurate for 3D lidar mapping, we just rely on detecting and measuring the possible ghosting in the point cloud map. However, in the actual application, we found that low bushes and some lush leaves would cause our algorithm to falsely detect ghosting points. Of course, this is taken for granted in the view of our basic principle of detecting ghosting as shown in Fig. \ref{ghdt}, because the same leaf or the same willow branch is really hard to remain stationary during multiple map collections and it can easily be blown by the wind to change its position in space. But this phenomenon will affect the precision rate of our algorithm in detecting bad poses.

Therefore, in addition to the simple semantic information of the ground and pole-liked points, more point cloud semantic information would be used to reduce the adverse impact of bushes or lush leaves on our proposed method.

\section{\textsc{Conclusions}}

In this paper, we proposed a novel method to automatically evaluate the relative accuracy for 3D lidar mapping without the ground truth of poses or map. By detecting and measuring the ghosting in the submap around the pose to be evaluated, we can indirectly judge whether it is a bad pose whose relative accuracy is less than 0.1m. Thus our method finally calculates the percentage $\mathcal{P}_{bad}$ of bad poses in all estimated poses and then outputs the common evaluation metric $\mathcal{P}_{acc} = 1 - \mathcal{P}_{bad}$ of the HD map for autonomous vehicles. We built four real-world benchmark map data to verify the precision and recall rate of detecting bad poses by our method, the experimental results demonstrate that the precision rate is greater than 98\% and the recall rate is 100\%. Moreover, our method can realize the 3D accuracy evaluation of poses by detecting the ghosting distributed along the three axes of XYZ, and the whole evaluation process is fully automated which greatly reduces the labor cost of quality inspection for HD map. The efficiency of our method in the hardware condition mentioned in this paper is up to $3.3min/km$, which can be further improved once the method is parallelized on a larger scale. In the future, we will use more point cloud semantic information to improve the precision rate and robustness of our method.
 
\addtolength{\textheight}{-12cm}   


\bibliographystyle{ieeetr}
\bibliography{references} 

\end{document}